\documentclass[journal]{IEEEtran}
\ifCLASSINFOpdf
\else
\fi

\usepackage{graphicx}
\usepackage{times}
\usepackage{epsfig}
\usepackage{graphicx}
\usepackage{amsmath}
\usepackage{amssymb}
\usepackage{makecell}
\usepackage{color, colortbl}
\definecolor{LightCyan}{rgb}{0.88,1,1}
\usepackage[shortlabels]{enumitem}

\begin{document}
%
\title{Action Recognition with Spatio-Temporal \\ Visual Attention on Skeleton Image Sequences}
%
%
%
        
\author{Zhengyuan~Yang,~\IEEEmembership{Student Member,~IEEE,}
        Yuncheng~Li, 
        Jianchao~Yang,~\IEEEmembership{Member,~IEEE,}
        and~Jiebo~Luo,~\IEEEmembership{Fellow,~IEEE}
}

%
%

\markboth{ }%
{Shell \MakeLowercase{\textit{et al.}}: Bare Demo of IEEEtran.cls for IEEE Journals}
%



\maketitle

\begin{abstract}
Action recognition with 3D skeleton sequences is becoming popular due to its speed and robustness. The recently proposed Convolutional Neural Networks (CNN) based methods have shown good performance in learning spatio-temporal representations for skeleton sequences. Despite the good recognition accuracy achieved by previous CNN based methods, there exist two problems that potentially limit the performance. First, previous skeleton representations are generated by chaining joints with a fixed order. The corresponding semantic meaning is unclear and the structural information among the joints is lost. Second, previous models do not have an ability to focus on informative joints. The attention mechanism is important for skeleton based action recognition because there exist spatio-temporal key stages while the joint predictions can be inaccurate. To solve these two problems, we propose a novel CNN based method for skeleton based action recognition. We first redesign the skeleton representations with a depth-first tree traversal order, which enhances the semantic meaning of skeleton images and better preserves the associated structural information. We then propose the idea of a two-branch attention architecture that focuses on spatio-temporal key stages and filters out unreliable joint predictions. A base attention model with the simplest structure is first introduced to illustrate the two-branch attention architecture. By improving the structures in both branches, we further propose a Global Long-sequence Attention Network (GLAN). Furthermore, in order to adjust the kernel's spatio-temporal aspect ratios and better capture long term dependencies, we propose a Sub-Sequence Attention Network (SSAN) that takes sub-image sequences as inputs. We show that the two-branch attention architecture can be combined with the SSAN to further improve the performance. Our experiment results on the NTU RGB+D dataset and the SBU Kinetic Interaction dataset outperforms the state-of-the-art. The model is further validated on noisy estimated poses from the UCF101 dataset and the Kinetics dataset.
\end{abstract}

\begin{IEEEkeywords}
Action and Activity Recognition, Video Understanding, Human Analysis, Visual Attention.
\end{IEEEkeywords}

%
\IEEEpeerreviewmaketitle

\section{Introduction}
%
%
%
%
\IEEEPARstart{T}{he} major modalities used for action recognition include RGB videos \cite{donahue2015long,tran2015learning,Qiu_2017_ICCV}, optic flow \cite{simonyan2014two,feichtenhofer2016convolutional,carreira2017quo} and skeleton sequences. Compared to RGB videos and optic flow, skeleton sequences are computationally efficient. Furthermore, skeleton sequences have a better ability to represent dataset-invariant action information since no background context is included. One limitation is that manually labeling skeleton sequences is too expensive, while  automatic annotation methods may yield inaccurate predictions. Given the above advantages and the fact that skeletons can now be more reliably predicted \cite{shotton2013real,mehta2017vnect}, skeleton based human action recognition is becoming increasingly popular. The major goal for skeleton based recognition is to learn a representation that best preserves the spatio-temporal relations among the joints.

With a strong ability of modeling sequential data, Recurrent Neural Networks (RNN) with Long Short-Term Memory (LSTM) neurons outperform the previous hand-crafted feature based methods \cite{yang2012eigenjoints,vemulapalli2014human}. Each skeleton frame is converted into a feature vector and the whole sequence is fed into the RNN.
Despite the strong ability in modeling temporal sequences, RNN structures lack the ability to efficiently learn the spatial relations between the joints. To better use spatial information,  a hierarchical structure is proposed in  \cite{du2015hierarchical,shahroudy2016ntu} that feeds the joints into the network as several pre-defined body part groups. However, the pre-defined body regions still limit the effectiveness of representing spatial relations.  A spatio-temporal 2D LSTM (ST-LSTM) network \cite{liu2016spatio} is proposed to learn the spatial and temporal relations simultaneously. Furthermore, a two-stream RNN structure \cite{wang2017modeling} is proposed to learn the spatio-temporal relations with two RNN branches.

CNN has a natural ability to learn representations from 2D arrays. The works in \cite{ke2017new,kim2017interpretable} first propose to represent the skeleton sequences as 2D gray scale images and use CNN to jointly learn a spatio-temporal representation. Each gray scale image corresponds to one axis in the joint coordinates. For example, the coordinates in the x-axis throughout a skeleton sequence generate one single-channel image. Each row is a spatial distribution of coordinates at a certain time-stamp, and each column is the temporal evolution of a certain joint. The generated 2D arrays are then scaled and resized into a fixed size. The gray scale images generated from the same skeleton sequence are concatenated together and processed as a multi-channel image, which is called the {\it skeleton image}.

Despite the large boost in recognition accuracy achieved by previous CNN based methods, there exist two problems. 
First, previous skeleton image representations lose spatial information. In previous methods, each row represents skeleton's spatial information by chaining all joints with a fixed order. This concatenation process lacks semantic meaning and leads to a loss in skeleton's structural information. Although a good chain order can perverse more spatial information, it is impossible to find a perfect chain order that maintains all spatial relations in the original skeleton structure. We propose a Tree Structure Skeleton Image (TSSI) to preserve spatial relations. TSSI is generated by traversing a skeleton tree with a depth-first order. We assume the spatial relations between joints are represented by the edges that connect them in the original skeleton structure, as shown in Figure \ref{fig:TSSI} (a). The fewer edges there are, the more relevant the joint pair is. Thus we prove that TSSI best preserves the spatial relation. 

Second, previous CNN based methods do not have the ability to focus on spatial or temporal key stages. In skeleton based action recognition, certain joints and frames are more informative, like the joints on the arms in action `waving hands'. Furthermore, certain joints may be inaccurately predicted and should be neglected. Therefore, it is important to include attention mechanisms. The skeleton image representation has a natural ability to represent spatio-temporal importance jointly with 2D attention masks, where each row represents the spatial importance of key joints and each column represents the temporal importance of key frames. We propose a two-branch architecture for visual attention on single skeleton images. One branch generates an attention mask with a larger receptive field and the other branch refines the CNN feature. We first introduce the two-branch architecture with a base attention model. Furthermore, a Global Long-sequence Attention Network (GLAN) is proposed with refined branch structures. Experiments on public datasets prove the effectiveness of the two improvements. The recognition accuracy is superior to the state-of-the-art methods. 

Despite the effectiveness of the two-branch attention structure, representing an entire sequence as one skeleton image lacks the ability to adjust kernels' spatio-temporal resolutions and learn the long-term dependencies. The original resolutions are determined by the number of joints and the length of the sequence. Furthermore, there is information loss with a sequence longer than the height of the skeleton image. Therefore, we represent the skeleton sequence as several overlapped sub skeleton images and propose a Sub-Sequence Attention Network (SSAN) based on the representation. Furthermore, we show that the GLAN can be combined with the SSAN to further improve the performance.

Our main contributions include the following:
\begin{itemize}[nosep]
\item We propose a Tree Structure Skeleton Image (TSSI) that better preserves the spatial relations in skeleton sequences. TSSI is based on a depth-first tree traversal order instead of direct concatenation. 
\item We propose a two-branch visual attention architecture for skeleton based action recognition. A Global Long-sequence Attention Network (GLAN) is introduced based on the proposed architecture.
\item We propose a Sub-Sequence Attention Network (SSAN) to adjust spatio-temporal aspect ratios and better learn long-term dependencies. We further show that the GLAN and the SSAN can be well combined.
\item We evaluate the model on both reliable Kinect recorded poses and noisy estimated poses, which prove the robustness with incomplete input pose sequences.
\end{itemize}

\section{Related Work}
Human action recognition from videos is an important area in computer vision. \cite{wang2011action,wang2013action,peng2014action,peng2016bag} design hand-crafted features including HOG \cite{dalal2005histograms}, HOF or MBH \cite{dalal2006human} and encode the features for the recognition task. With the developments of deep neural networks \cite{krizhevsky2012imagenet,simonyan2014very}, more deep models are designed to solve the action recognition problem. Adopting ConvNets to encode frame-level feature and using LSTM to learn the temporal evolution has been proved to be an effective approach \cite{donahue2015long,srivastava2015unsupervised,yue2015beyond}. Optical flow \cite{simonyan2014two,feichtenhofer2016convolutional,carreira2017quo} is another way of representing temporal information. Furthermore, C3D \cite{tran2015learning} is proposed to learn the spatial and temporal information simultaneously with 3D convolutional kernels. A recent work in \cite{Qiu_2017_ICCV} can simulate 3D kernels. 

Compared to other frequently used modalities including RGB videos \cite{donahue2015long,tran2015learning,Qiu_2017_ICCV} and optical flow \cite{simonyan2014two,feichtenhofer2016convolutional,carreira2017quo}, skeleton sequences require much less computation and are more robust across views and datasets. With the advanced methods to acquire reliable skeletons from RGBD sensors \cite{shotton2013real} or even a single RGB camera \cite{mehta2017vnect, wei2016cpm, cao2016realtime,guler2018densepose}, skeleton-based action recognition is becoming increasingly popular.

Many previous skeleton-based action recognition methods \cite{lee2017ensemble} model the temporal pattern of skeleton sequences with Recurrent Neural Networks. Hierarchical structures \cite{du2015hierarchical,shahroudy2016ntu} better represent the spatial relations between body parts. Other works \cite{liu2017global,song2017end} adopt attention mechanisms to locate spatial key joints and temporal key stages in skeleton sequences. \cite{liu2016spatio} proposes a 2D LSTM network to learn spatial and temporal relations simultaneously. \cite{wang2017modeling} models spatio-temporal relations with a two-stream RNN structure. Other effective approaches include lie groups \cite{vemulapalli2014human,huang2016deep} and nearest neighbor search \cite{wengspatio}. Recently, graphical neural networks \cite{li2017action,yan2018spatial} achieve the state-of-the-art performance on the skeleton based recognition.

Compared to LSTM or graphical model based methods, the recently proposed CNN based approaches show a better performance in learning skeleton representations. \cite{ke2017new,kim2017interpretable} propose to convert human skeleton sequences into gray scale images, where the joint coordinates are represented by the intensity of pixels. \cite{liu2017skepxels} proposes to generate skeleton images with `Skepxels' to better represent the joint correlations. In this paper, we further improve the design of skeleton images with a depth-first traversal on skeleton trees.

Attention mechanisms are important for skeleton based action recognition. Previous LSTM based methods \cite{liu2017global,song2017end} learn attention weights between the stacked LSTM layers. For CNN based methods, we propose that general visual attention can be directly adopted to generate 2D attention masks, where each row represents spatial importance and each column represents temporal importance. Visual attention has achieved successes in many areas, including image captioning \cite{xu2015show,you2016image}, RGB based action recognition \cite{sharma2015action,du2017rpan}, image classification \cite{zheng2017learning,wang2017residual}, sentiment analysis \cite{you2017visual} and etc. Many visual attention methods take an image sequence as input \cite{sharma2015action,kim2017interpretable_car}, or use extra information from another modality like text \cite{xu2015show,you2016image,du2017rpan}. Because a single skeleton image already represents a spatio-temporal sequence without the need for an extra modality, we propose a single frame based visual attention structure with a same setting in \cite{zheng2017learning,wang2017residual}.

\section{Method}
In this section, we first introduce the previous design of skeleton images and the base CNN structure, before an improved Tree Structure Skeleton Image (TSSI) is proposed. Later, we propose the idea of two-branch visual attention and introduce a Global Long-sequence Attention Network (GLAN) based on the idea. Finally, we introduce the Sub-Sequence Attention Network (SSAN) to learn long-term dependencies.

\begin{figure}[t]
\begin{center}
   \centerline{\includegraphics[width=8cm]{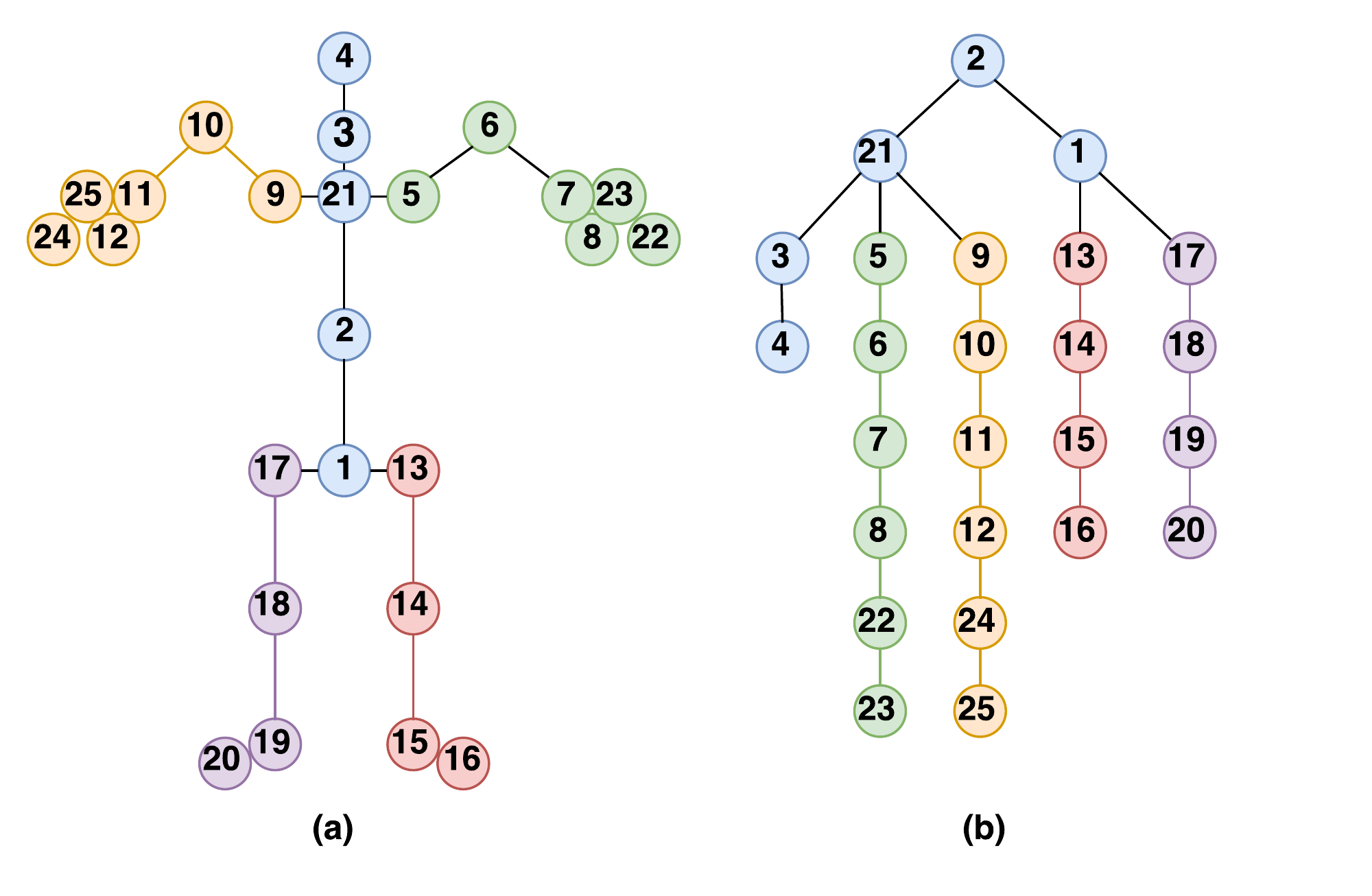}}
	\vspace{-0.1in}
   \centerline{\includegraphics[width=8cm]{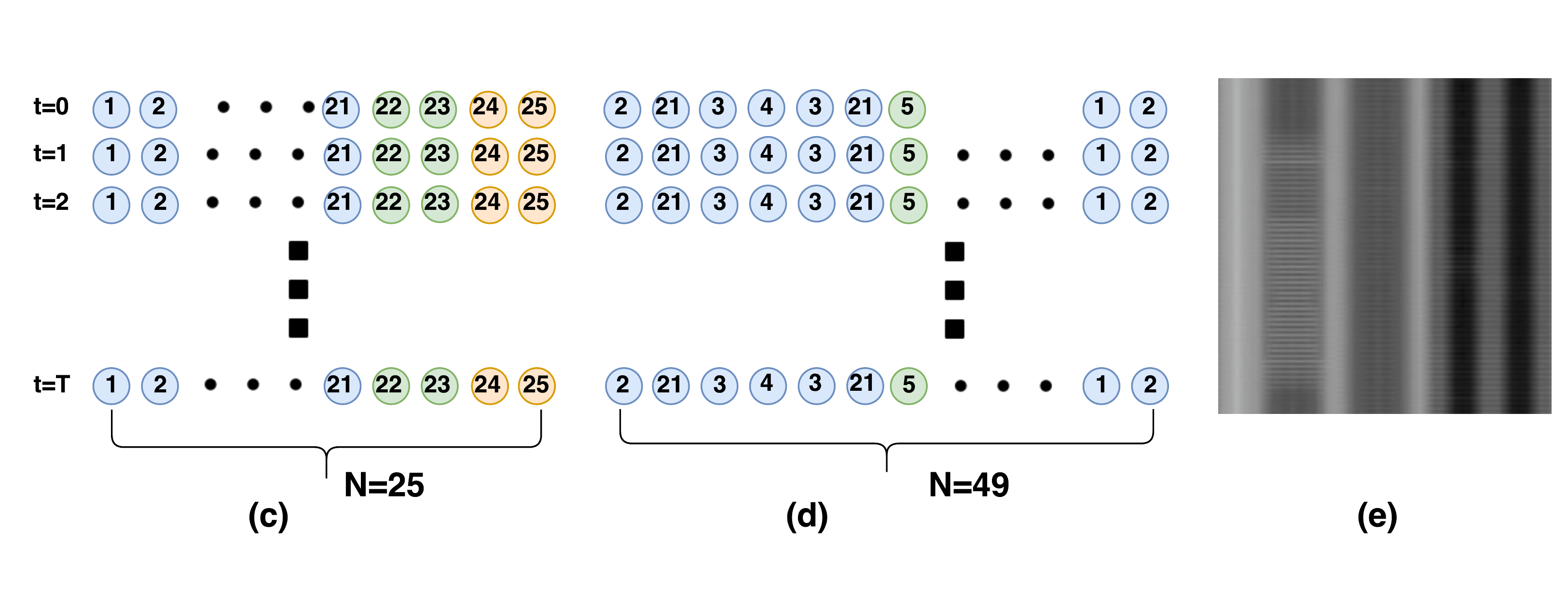}}
\end{center}
\vspace{-0.3in}
	\caption{Tree Structure Skeleton Image (TSSI): (a) Skeleton structure and order in NTU RGB+D, (b) Skeleton tree for TSSI generating, (c) Joint arrangements of naive skeleton images, (d) Joint arrangements of TSSI, and (e) An example frame of TSSI. Different colors represent different body parts.}
\label{fig:TSSI}
\end{figure}
\subsection{Base Model}
In CNN based skeleton action recognition, joint sequences are arranged as 2D arrays that are processed as gray scale images. We call such a generated image the `Skeleton Image'. For a channel in skeleton images, each row contains the chaining of joint coordinates at a certain time-stamp. Each column represents the coordinates of a certain joint throughout the entire video clip. The chain order of joints is pre-defined and fixed. A typical arrangement of the 2D array is shown in Figure \ref{fig:TSSI} (c). The generated 2D arrays are then scaled into 0 to 255, and resized into a fixed size of $224*224$. The processed 2D arrays are processed as gray scale images, where each channel represents an axis of joint coordinates. The skeleton images are fed into CNNs for action recognition. We use ResNet-50 \cite{he2016deep} as the base ConvNet model. Compared to RNN based or graph neural network based method, CNN based methods can better learn the spatio-temporal relations between joints. 


\subsection{Tree Structure Skeleton Image}
A shortcoming in the previous skeleton images is that each row is arranged by simply concatenating all joints. Each row contains the concatenation of all joints with a pre-defined chain order. CNN has a feature that the receptive field grows larger at higher levels. Therefore, the adjacent joints in each row or column are learned first at lower levels. This implies that the adjacent joints share more spatial relations in original skeleton structure, which often do not hold in previous skeleton images. 
In previous skeleton images, a generated array has 25 columns representing the joint coordinates of joint 1 to 25 with joint indexes shown in Figure \ref{fig:TSSI} (a). An arrangement of the skeleton image is shown in Figure \ref{fig:TSSI} (c). In this case, a convolutional kernel might cover joints [20, 21, 22, 23, 24] at a certain level since these joints are adjacent in skeleton images. However, these joints have less spatial relations in original skeleton structures and should not be learned together directly. 

To solve this problem, we propose a Tree Structure Skeleton Image (TSSI) inspired by \cite{liu2016spatio}. The basic assumption is that the spatially related joints in original skeletons have direct graph links between them. The less edges required to connect a pair of joints, the more related is the pair. The human structure graph is defined with semantic meanings as shown in \ref{fig:TSSI} (a). In the proposed TSSI, the direct concatenation of joints is replaced by a depth-first tree traversal order. The skeleton tree is defined in Figure \ref{fig:TSSI} (b) and an arrangement of TSSI is shown in Figure \ref{fig:TSSI} (d). The depth-first tree traversal order for each row is [2, 21, 3, 4, 3, 21, 5, 6, 7, 8, 22, 23, 22, 8, 7, 6, 5, 21, 9, 10, 11, 12, 24, 25, 24, 12, 11, 10, 9, 21, 2, 1, 13, 14, 15, 16, 15, 14, 13, 1, 17, 18, 19, 20, 19, 18, 17, 1, 2]. With the proposed order, the neighboring columns in skeleton images are spatially related in original skeleton structures. This proves that the TSSI best preserves the spatial relations. With TSSI, the spatial relations between related joints are learned first at lower levels of CNN and the relations between less relevant joints are learned later at high levels when receptive field becomes larger. An example of the generated TSSI is shown in Figure \ref{fig:TSSI} (e).

\subsection{Attention Networks}
In skeleton sequences, certain joints and frames are particularly distinguishable and informative for recognizing actions. For example in action `waving hands', the joints in arms are more informative. These informative joints and frames are referred to as `key stages'. Furthermore, noise exists in the captured joint data and deteriorates the recognition accuracy. The inaccurate joints should be automatically filtered out or ignored by the network. 

To alleviate the effect of data noises and to focus on informative stages, skeleton based methods should adjust weights for different inputs automatically. We propose the idea of two-branch visual attention and further design a Global Long-sequence Attention Network (GLAN) based on the idea. In this section, we first introduce the basic idea of the two-branch attention architecture with a base attention model. Then the structure of Global Long-sequence Attention Network (GLAN) is introduced.


\begin{figure}[t]
\begin{center}
   \centerline{\includegraphics[width=8cm]{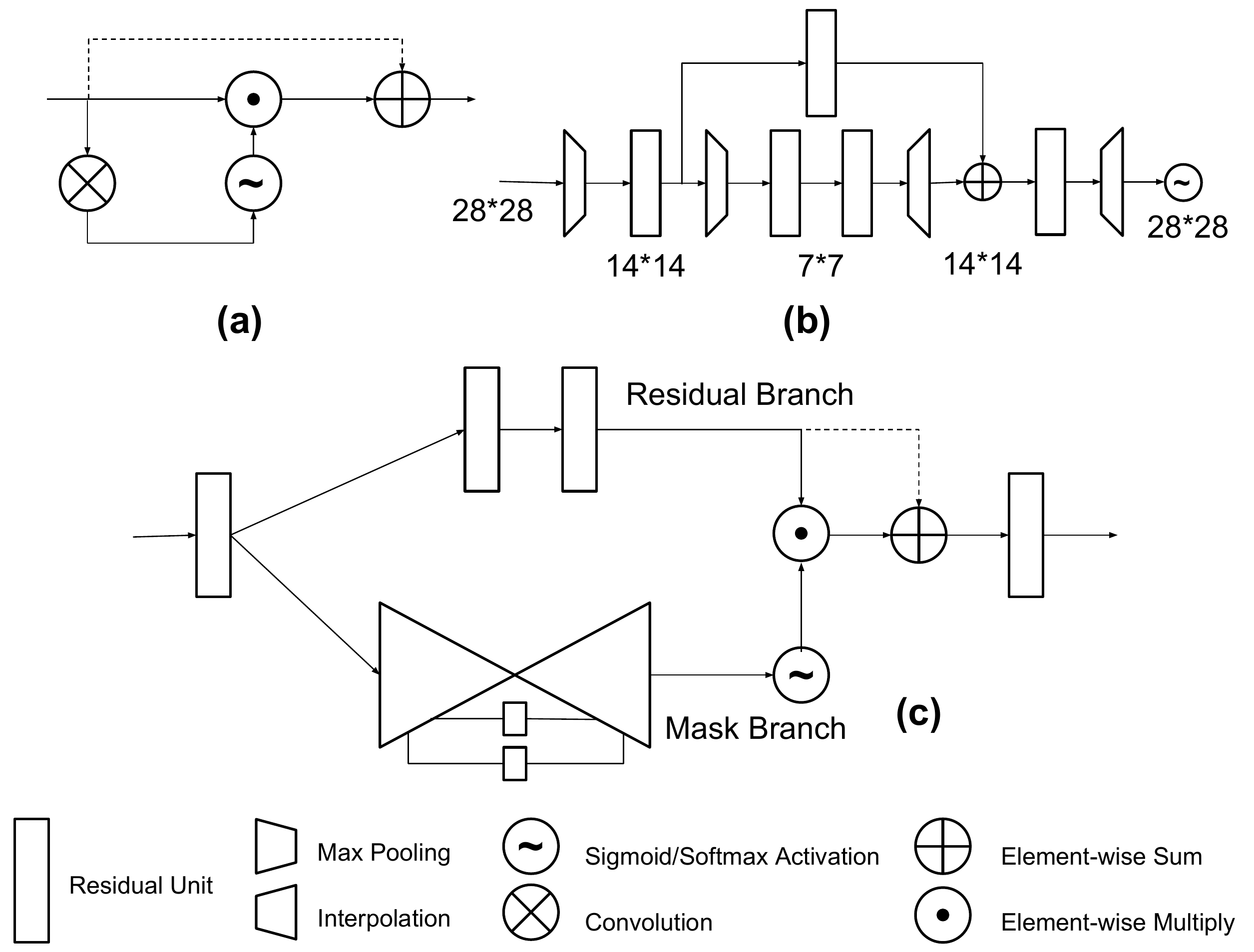}}
\end{center}
\vspace{-0.1in}
	\caption{A base attention module and a GLAN module: (a) A base attention block, (b) An expanded plot for the Hourglass mask branch in GLAN, (c) An attention block with GLAN structure, short for `GLAN block'.}
\vspace{-0.1in}
\label{fig:Weighting_module}
\end{figure}
{\bf Base Attention Model.} Skeleton images naturally represent both spatial and temporal information of skeleton sequences. Therefore a 2D attention mask can represent spatio-temporal importance simultaneously, where the weights in each row represent the spatial importance of joints and the weight in each column represent the temporal importance of frames. In order to generate the attention masks, we propose a two-branch attention architecture that learns attention masks from a single skeleton image. 
The two-branch structure consists of `mask branches' and `residual branches'. Taking previous CNN feature blocks as inputs, the mask branch learns a 2D attention mask and the residual branch refines previous CNN feature. The two branches are then merged to output a weighted CNN feature block. To be specific, the mask branch learns an attention mask with a structure that has a larger receptive field. The residual branch is designed to maintain and refine the input CNN features with convolutional layers. The two branches are fused at the end of each attention block with element-wise multiplication and summation. 

\begin{figure*}[t]
\begin{center}
   \centerline{\includegraphics[width=16cm,height=5cm]{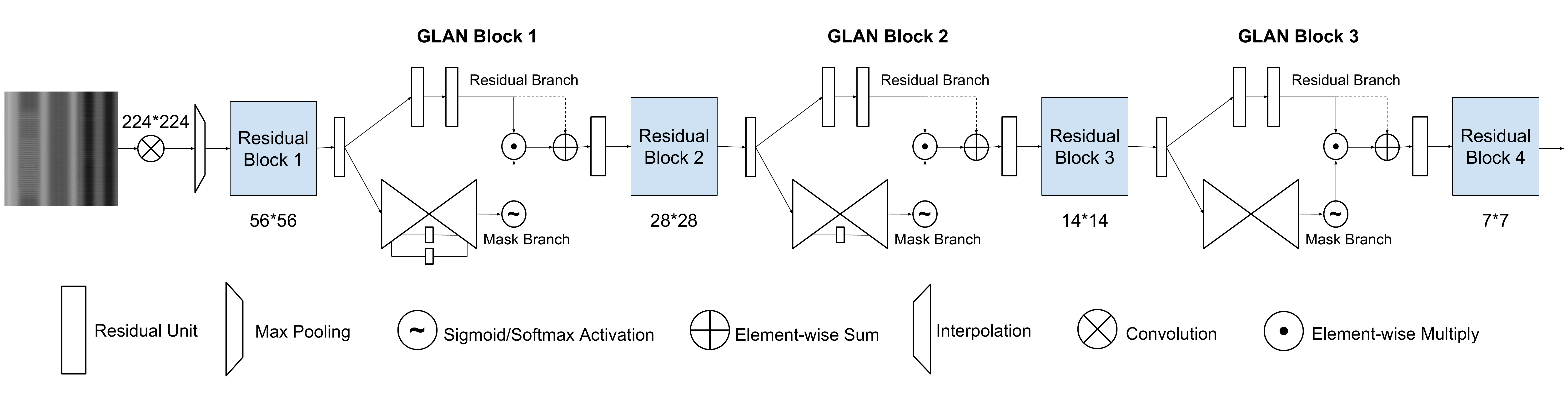}}
\end{center}
\vspace{-0.3in}
	\caption{The framework for Global Long-sequence Attention Network (GLAN).}
\label{fig:GLAN_ALL}
\end{figure*}

We first introduce the base attention model, which is the simplest version of two-branch attention structures. As shown in Figure \ref{fig:Weighting_module} (a), the mask branch in the base model gains a larger receptive field with a single convolutional layer. Softmax or Sigmoid functions are used for mask generating. The residual branch preserves the input CNN feature with a direct link. An `attention block' is defined as a structure with one mask branch and one residual branch as Figure \ref{fig:Weighting_module} (a). Attention blocks are added between the convolutional blocks in base CNN to build the whole network. In the base attention model, attention blocks are inserted between ResNet-50's residual blocks, with the structure of residual blocks unchanged.



{\bf Global Long-sequence Attention Network (GLAN).}
Based on the proposed two-branch structure, we improve the designs of both branches to learn masks and CNN features more effectively. Inspired by the hourglass structure \cite{newell2016stacked,wang2017residual}, we propose a Global Long-sequence Attention Network (GLAN) as shown in Figure \ref{fig:GLAN_ALL}. The hourglass structure is adopted in mask branches to quickly adjust the feature size and efficiently gain a larger receptive field. As shown in Figure \ref{fig:Weighting_module} (b), the hourglass structure consist of a series of down-sampling units followed by up-sampling units.
In each hourglass mask branch, input CNN features are first down-sampled to the lowest spatial resolution of $7*7$ and recovered back to the original size. Max pooling is used for down-sampling and bilinear interpolation is used for up-sampling. Each down-sampling unit includes a max pooling layer, a followed residual unit and a link connection to the recovered feature with a same size. Each up-sampling unit contains a bilinear interpolation layer, a residual unit and a element-wise sum with the link connection. We show that the Convolution-Deconvolution structure gains a large receptive field effectively and therefore can better learn an attention mask. For residual branches, we add two residual units to further refine the learned CNN feature. All residual units are the same as ResNet-50 \cite{he2016deep}, which contains three convolutional units and a direct residual link.

As shown in Figure \ref{fig:GLAN_ALL}, three GLAN attention blocks are added between the four residual blocks in ResNet-50 to build the GLAN network. The depth of each GLAN blocks varies due to the different input feature sizes. Furthermore, we reduce the number of residual units in each residual block to keep a proper depth of the GLAN network, since GLAN blocks are much deeper than the base attention blocks. Only one residual unit is kept for the first three residual blocks. The final residual block keeps all three residual units as in ResNet-50.


\subsection{Long-term Dependency Model}

Although single-frame based two-branch attention structure has a good performance, it lacks the ability to learn long-term dependences. The generated skeleton image has a fix height of 224. This implies an information loss with a sequence longer than 224, which is around 7 seconds with a frame rate of 30 fps. To better learn long-term dependencies, we propose a CNN + LSTM model with sub skeleton image sequences. We first split skeleton sequences into several overlapped sub-sequences and generate a series of sub skeleton images for a skeleton sequence. CNN features are first extracted from each sub-sequence skeleton image and the long-term dependencies are modeled with RNNs. 

Furthermore, in the original two-branch attention structures, both the spatial and temporal resolutions in skeleton images are fixed by the number of joints and the length of the sequence. However, the kernel should be able to adjust the number of joints and frames it looks at jointly to achieve the best performance. The proposed sub-image model can flexibly adjust the relative resolution by adjusting the number of sub-images and the overlapping rate. This adjustment is equivalent to adjusting the width and height of CNN kernels, while it does not need to retrain the model every time.

\begin{figure}[t]
\begin{center}
   \centerline{\includegraphics[width=9cm]{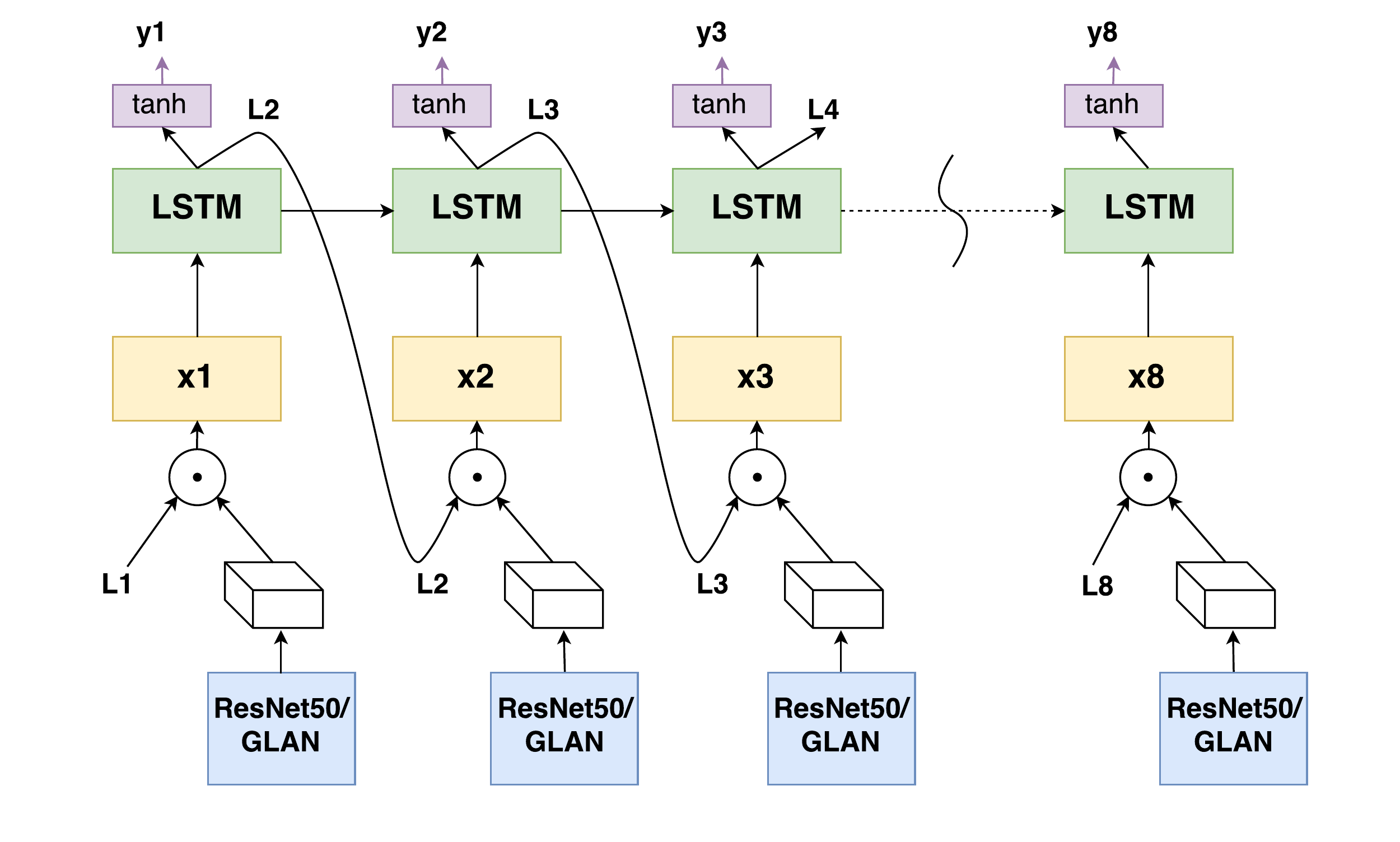}}
\end{center}
\vspace{-0.25in}
	\caption{Sub-Sequence Attention Network (SSAN).}
\label{fig:SSAN}
\end{figure}
{\bf Sub-Sequence Attention Network (SSAN).} Based on the proposed sub-image model, an long-term attention module is added. Inspired by \cite{sharma2015action,xu2015show}, a Sub-Sequence Attention Network (SSAN) is proposed with a structure shown in Figure \ref{fig:SSAN}.
We adopt Long Short-Term Memory (LSTM) as the RNN cell. The LSTM implementation is based \cite{zaremba2014recurrent,xu2015show}:
\begin{equation}
\left(
  \begin{matrix}
  i_t\\f_t\\o_t\\g_t
  \end{matrix}
\right)=
\left(
  \begin{matrix}
  \sigma\\\sigma\\\sigma\\\tanh
  \end{matrix}
\right)T_{d+D,4d}
\left(
  \begin{matrix}
  h_{t-1}\\x_t
  \end{matrix}
\right)
\end{equation}
\begin{equation}
c_t=f_t\odot c_{t-1}+i_t\odot g_t
\end{equation}
\begin{equation}
h_t=o_t\odot \tanh(c_t)
\end{equation}
where $i_t,f_t,o_t,c_t,h_t$ are the input, forget, output, memory and hidden states for the LSTM. $T$ is an affine transformation, where $D$ is the depth of the CNN feature block and $d$ is the dimension of all LSTM states. $x_t$ is the CNN feature input to the LSTM at time $t$ with length $D$.

Based on the LSTM model, the 2D attention map at time $t$ $l_{t}$ is defined as a $K*K$ mask, where $K$ is the output width and height of the CNN feature block:
\begin{equation}
l_{t,i}=\frac{\exp(W_i^\top h_{t-1})}{\sum_{j=1}^{K*K} \exp(W_j^\top h_{t-1})} \quad i \in 1 \dots K^2
\end{equation}
Inspired by \cite{wang2017residual}, we also adopt the spatial-channel attention with sigmoid activation, where $i \in 1 \dots K^2, z \in 1 \dots D$.
\begin{equation}
l_{t,i,z}=Sigmoid(W_{i,z}^\top h_{t-1})
\end{equation}
The weighted CNN feature at time $t$ $x_t$ is the element-wise multiplication of attention mask $l_t$ and original CNN output $X_t$ following Equation \ref{equ:weighted_x}. In the SSAN, Resnet-50 is selected as the CNN model.
\begin{equation}
x_t=\sum_{j=1}^{K^2} l_{t,i} X_{t,i}
\label{equ:weighted_x}
\end{equation}

{\bf GLAN + SSAN.} Furthermore, we show that the GLAN can replace Resnet-50 as the CNN structure in the long-term dependency model. The combination of SSAN and GLAN enables the framework to generate attentions both early in CNN layers with a bottom-up approach and long-term in LSTM layers with a top-down approach. Experiments show the effectiveness of the proposed combination, and further prove the possibility of using the proposed module as atomic parts in other frameworks.

\section{Experiments}
The proposed method is evaluated on both  clean datasets captured by Kinect and noisy datasets where the poses are estimated from RGB videos. We adopt the NTU RGB+D dataset \cite{shahroudy2016ntu} and the SBU Kinect Interaction Dataset \cite{yun2012two} for clean dataset evaluation. The estimated poses on UCF101 \cite{soomro2012ucf101} and Kinetics \cite{kay2017kinetics} are used to measure the performance with potentially incomplete and noisy poses.
We further evaluate the effectiveness of each proposed module separately. The experiments show that both the TSSI and the attention network generate a large boost in the action recognition accuracy to outperform the state-of-the-art.

\subsection{Datasets}
{\bf NTU RGB+D.} The NTU RGB+D dataset \cite{shahroudy2016ntu} is so far the largest 3D skeleton action recognition dataset. NTU RGB+D has 56880 videos collected from 60 action classes, including 40 daily actions, 9 health-related actions and 11 mutual actions. The dataset is collected with Kinect and the recorded skeletons include 25 joints. The train/val/test split follows  \cite{shahroudy2016ntu}. Samples with missing joints are discarded as in that paper.

{\bf SBU Kinect Interaction.} The SBU Kinect Interaction dataset \cite{yun2012two} contains 282 skeleton sequences and 6822 frames. We follow the standard experiment protocol of 5-fold cross validations with the provided splits. The dataset contains eight classes. There are two persons in each skeleton frame and 15 joints are labeled for each person. The two skeletons are processed as two data samples during training and the averaged prediction score is calculated for testing. 

{\bf UCF101.} The UCF101 dataset \cite{soomro2012ucf101} contains 13,320 videos from 101 action categories. Videos have a fixed frame rate and resolution of 25 FPS and $320\times240$. Based on the RGB videos, we estimate the 18 joints poses with a state-of-the-art pose estimation toolbox, OpenPose \cite{cao2016realtime}. The toolbox provides 2D joint locations and the confidence value for the prediction. Furthermore, we include pre-processing steps to alleviate the problem of incomplete poses, which will be discussed in Section \ref{noisy_results}. We will release both the estimated and processed poses on UCF101 for fair comparison.

{\bf Kinetics.} The Kinetics dataset \cite{kay2017kinetics} is the largest RGB action recognition dataset, containing around 300,000 video clips from 400 action classes. The videos are collected from YouTube and each clips is around 10 seconds long. Similarly, OpenPose is used for pose estimation. We use the pre-calculated estimated poses provided by \cite{yan2018spatial}, which converts the videos into a fixed resolution of $340\times256$ and a frame rate of 30 FPS. Furthermore, the model is evaluated on the `Kinetics-Motion' dataset \cite{yan2018spatial}, which is a 30 classes subset of Kinetics with action labels strongly related to body motions. The label names are: {\it belly dancing, punching bag, capoeira, squat, windsurfing, skipping rope, swimming backstroke, hammer throw, throwing discus, tobogganing, hopscotch, hitting baseball, roller skating, arm wrestling, snatch weight lifting, tai chi, riding mechanical bull, salsa dancing, hurling (sport), lunge, skateboarding, country line dancing, juggling balls, surfing crowd, dead lifting, clean and jerk, crawling baby, push up, front raises, pull ups}.


\subsection{Ablation Studies}
To prove the effectiveness of the TSSI and the proposed attention networks, we separately evaluate each proposed module with results shown in Table \ref{table:NTU_state}. Each component of the framework is evaluated on NTU RGB+D with the cross subject setting. NTU RGB+D is selected for component evaluations because it is the largest and the most challenging dataset so far. Similar results are observed on other datasets.

{\bf Traditional Skeleton Image + ConvNet.} As a baseline, we adopt the previous skeleton image representation from \cite{ke2017new} and use ResNet-50 as a base CNN model to train spatio-temporal skeleton representations. We test the three spatial joint orders proposed by Sub-JHMDB \cite{jhuang2013towards}, PennAction \cite{zhang2013actemes} and NTU RGB+D \cite{shahroudy2016ntu}. Experiments show that the NTU RGB+D's order generates a better accuracy of $1.3\%$ than the rest two orders. Therefore, we adopt the joint order proposed by NTU RGB+D for baseline comparison. The order is shown in Figure \ref{fig:TSSI} (a). 

{\bf TSSI + ConvNet.} The effectiveness of the proposed Tree Structure Skeleton Image (TSSI) is compared to the baseline design of skeleton images. TSSI is the skeleton image generated with a depth-first tree traversal order. The skeleton tree structure, TSSI arrangement and a TSSI example is shown in Figure \ref{fig:TSSI} (b), (d), (e). A large boost in accuracy is observed from $68.0\%$ to $73.1\%$, which proves the effectiveness of TSSI.

{\bf TSSI + Base Attention.} The base attention model provides a baseline for two-branch attention networks. 
The base attention blocks with and without residual links are inserted at three different locations in ResNet-50, that is at the front after the first convolutional layer, in the middle after the second residual block and in the end after the final residual block. The input feature blocks to the three attention blocks have the shapes of $112*112*64$, $28*28*512$ and $7*7*2048$. The recognition accuracy boosts from $73.1\%$ to $74.9\%$. This experiment shows that even the simplest two-branch attention network can improve the recognition accuracy.

{\bf TSSI + GLAN.} We evaluate the proposed Global Long-sequence Attention Network (GLAN). The number of link connections and the depth of the hourglass mask branch can be manually adjusted. In experiments, we first down-sample the feature blocks to a lowest resolution of $7*7$ and then up-sample them back to the input size. Each max pooling layer goes with one residual unit, one link connection and one up-sampling unit. With a GLAN structure shown in Figure \ref{fig:GLAN_ALL}, the recognition accuracy increases from $74.9\%$ to $80.1\%$.

{\bf TSSI + SSAN.} The SSAN is one of the two attention networks we proposed. The number of sub-sequences and the overlapping rate for the sub-sequences are two hyper-parameters that are tuned with validation set. With a sub-sequence number of 5 and an overlapping rate of 0.5, the attention network achieves an accuracy of $80.9\%$ from $73.1\%$.

\begin{table}
\caption{GLAN + SSAN performance with different hyper parameters on the NTU RGB+D dataset.}
\vspace{-0.15in}
\begin{center}
\begin{tabular}{ |c|c|c|c|}
    \hline
    \thead{Sub-image\\Lengths} & \thead{Sub-image\\Numbers} & \thead{Overlapping\\Rate} & Accuracy (\%)\\
    \hline
    $T/3$ & 3 & 0\% & 80.80 \\
    \rowcolor{LightCyan}
    $T/3$ & 5 & 50\% & 82.42 \\
    $T/3$ & 9 & 75\% & 81.38 \\
    \hline
    $T/5$ & 5 & 0\% & 78.56 \\
    $T/4$ & 5 & 25\% & 80.30 \\
    \rowcolor{LightCyan}
    $T/3$ & 5 & 50\% & 82.42 \\
    $T/2$ & 5 & 75\% & 81.57 \\
    \hline
\end{tabular}
\end{center}
\vspace{-0.18in}
\label{table:hyper_para}
\end{table}

{\bf TSSI + GLAN + SSAN.} Finally, we show that the GLAN and SSAN can be well combined to further improve the performance. By replacing the Resnet-50 with the proposed GLAN, the framework achieved an accuracy of $82.4\%$. This experiment also shows that the proposed two branch attention structure can be adopted as atomic CNN structure in various frameworks to achieve a better performance.

Furthermore, we analyze the hyper-parameters in the SSAN, i.e. the overlapping rate, the number and length of sub-images. The relation of these parameters is: 
\begin{equation}
T=t_{sub}*[1+(1-\alpha)*(n-1)]
\end{equation}
where $t_{sub}$ is the number of frames in each sub-image or the sub-image length, $T$ is the number of frames in the whole sequence, $\alpha$ is the overlapping rate and $n$ is the number of sub-images. We design two sets of experiments with fixed sub-image lengths or fixed sub-image numbers to interpret the effectiveness of the SSAN and find the best set of hyper parameters. Experiments are conducted on NTU RGB+D with the TSSI + GLAN + SSAN framework. 

Starting from the optimal hyper-parameters of a $50\%$ overlapping rate and 5 sub-images, we report the performances under different hyper parameters with either sub-image numbers or sub-image lengths unchanged. For the fixed length experiment as shown in Table \ref{table:hyper_para}, the length of sub-images are fixed and the number of sub-images changes from 3 to 9 by adjusting the overlapping rate. We observe the accuracy drops $1.6\%$ from $82.4\%$ to $80.8\%$. In the fixed sub-image number experiment, the number of sub-images is fixed as five where the best performance is achieved. The length of sub-images varies from $T/5$ to $T/2$ with different overlapping rates. A larger drop of accuracy of $3.8\%$ is observed in the fixed sub-image number experiment.

According to the experiment results, the length of sub-images influence the performance of the SSAN most. This implies that the SSAN produces a large boost in accuracy mainly with its ability to flexibly adjusting the spatial-temporal resolutions. The optimal hyper parameters of a $50\%$ overlapping rate, 5 sub-images and the $T/3$ temporal length works best  with the 25 joints on NTU RGB-D. Different temporal lengths can be adjusted for different numbers of annotated joints. Furthermore, the SSAN also better learns the long term dependencies that most results with the SSAN as shown in Table \ref{table:hyper_para} outperform the TSSI + GLAN.

\begin{table}
\caption{The action recognition accuracy compared to the state-of-the-art methods on the NTU RGB+D dataset.}
\vspace{-0.15in}
\begin{center}
\begin{tabular}{ c c c }
    \hline
    State-of-the-art & \thead{Cross\\Subject} & \thead{Cross\\View}\\
    \hline
	Lie Group \cite{vemulapalli2014human} & 51.0 & 52.8\\
	HBRNN \cite{du2015hierarchical} & 59.1 & 64.0\\
	Part-aware LSTM \cite{shahroudy2016ntu} & 62.9 & 70.3\\
	Trust Gate LSTM \cite{liu2016spatio} & 69.2 & 77.7 \\
	Two-stream RNN \cite{wang2017modeling} & 71.3 & 79.5 \\
	TCN \cite{kim2017interpretable} & 74.3 & 83.1 \\
	Global Attention LSTM \cite{liu2017global} & 74.4 & 82.8 \\
	$\mathrm{A^2GNN}$ \cite{li2017action} & 72.7 & 82.8 \\
	Clips+CNN+MTLN \cite{ke2017new} & 79.6 & 84.8 \\
	Ensemble TS-LSTM \cite{lee2017ensemble} & 76.0 & 82.6 \\
	Skepxels \cite{liu2017skepxels} & 81.3 & {\bf 89.2} \\
	ST-GCN \cite{yan2018spatial} & 81.5 & 88.3 \\
    \hline
	Proposed Model & \thead{Cross\\Subject} & \thead{Cross\\View} \\
    \hline
	Base Model & 68.0 & 75.5\\
	With TSSI & 73.1 & 76.5\\
	TSSI + Base Attention & 74.9 & 79.1\\
	TSSI + GLAN \cite{yang2018action} & 80.1 & 85.2 \\
	TSSI + SSAN & 80.9 & 86.1\\
	TSSI + GLAN + SSAN & {\bf 82.4} & 89.1\\
\end{tabular}
\end{center}
\vspace{-0.18in}
\label{table:NTU_state}
\end{table}

\subsection{Evaluations on Clean Datasets}
{\bf NTU RGB+D.} 
The middle column of Table \ref{table:NTU_state} shows the results of the NTU RGB+D cross subject setting. The base model with naive skeleton images already outperforms a number of previous LSTM based method, without adopting the attention mechanism. This shows that CNN based methods are promising for skeleton based action recognition. With the improved TSSI, the cross subject accuracy achieves $73.1\%$, which is comparable to the state-of-the-art LSTM methods. The proposed two-branch attention architecture further improves the performance and the GLAN outperforms the state-of-the-art. Experiments prove the effectiveness of the proposed CNN based action recognition method.

Furthermore, we show that generating sub-sequences and adopting the long-term dependency model (SSAN) can achieve a better results. The SSAN with ResNet-50 achieves a cross subject accuracy of $80.9\%$. By replacing the ResNet with the proposed GLAN to provide the spatial attention, the framework improves the state-of-the-art to $82.4\%$.

Similar results are observed in the NTU RGB+D cross view setting, as shown in the right column of Table \ref{table:NTU_state}. 

{\bf SBU Kinect Interaction.}
Similar to the performance on the NTU RGB+D dataset, the proposed TSSI and attention framework generates a large boost in the recognition accuracy on the SBU Kinect Interaction dataset that outperforms the state-of-the-art. The performances are shown in Table \ref{table:SBU_state}.

\begin{table}
\caption{The recognition accuracy compared to the state-of-the-art methods on the SBU Kinetic Interaction dataset.} 
\vspace{-0.15in}
\begin{center}
\begin{tabular}{ c c }
    \hline
    State-of-the-art & Accuracy (\%)\\
    \hline
	Raw Skeleton \cite{yun2012two} & 49.7 \\
	HBRNN \cite{du2015hierarchical} & 80.4 \\
	Trust Gate LSTM \cite{liu2016spatio} & 93.3 \\
	Two-stream RNN \cite{wang2017modeling} & 94.8 \\
	Global Attention LSTM \cite{liu2017global} & 94.1 \\
	Clips+CNN+MTLN \cite{ke2017new} & 93.6 \\
    \hline
	Proposed Model & Accuracy (\%) \\
    \hline
	Base Model & 82.0\\
	With TSSI & 89.2\\
	TSSI + Base Attention & 93.6\\
	TSSI + GLAN \cite{yang2018action} & 95.4\\
	TSSI + SSAN & 94.0\\
	TSSI + SSAN + GLAN & {\bf 95.7}\\
\end{tabular}
\end{center}
\vspace{-0.18in}
\label{table:SBU_state}
\end{table}

\subsection{Error Case Analysis}
To better understand the successful and failure cases, experiments are conducted to analyze the performance of each class in NTU RGB+D. As shown in Table \ref{table:error_case}, two parts of analysis are conducted. First, eight classes that constantly perform the best or worst are selected on the left side of Table \ref{table:error_case}. Results show that the actions with dynamic body movements, such as standing, sitting and walking, can be well classified with skeletons, while the classes with less motions like reading, writing and clapping usually have a poor result. The first, middle and last frames from these classes are visualized in the first row of Figure \ref{fig:example_frame}. This follows human intuition that skeletons are more useful for distinguishing dynamic actions, while additional background context information is necessary for recognizing the actions with less motions. 

The results also show that the proposed TSSI, GLAN and SSAN all generate a large boost in performance in all the listed classes. On the righthand side of the table, statistics of the best and worst classes are listed.
Results show that TSSI + GLAN + SSAN greatly improve the accuracy in challenging classes. The top 1 worst class in TSSI + GLAN + SSAN has an accuracy of $42.8\%$, which is even better than the averaged accuracy of the worst 10 in base model. For the best classes, the top 1 accuracy between the baseline and TSSI + GLAN + SSAN are similar. The improvements are mainly obtained through the improvements in the challenging classes. 


\subsection{Self-Paced Learning on Noisy Datasets}
\label{noisy_results}
The model is then evaluated on large scale RGB datasets with estimated and thus noisy poses. For a fair comparison, we do not use any pre-processing methods like interpolation to reduce the noise. 
To better learn a noise robust system, we adopt self-paced learning \cite{luo2001self,kumar2010self,jiang2014easy,jiang2014self} during the training process. The model is first trained with a small portion of reliable pose estimations and then gradually take more noisy data as inputs.
The average pose estimation confidence values provided by Openpose is used as the indication of reliability and the level of noises. The model starts with a high confidence threshold of $0.5$, i.e. all estimated pose sequences with an average confidence lower than $0.5$ are eliminated in the training process. We then fine-tune the model step by step by feeding more unreliable noisy data. Experiments show that self-paced learning can both accelerate the convergence speed and improve the final accuracy.

\begin{figure*}[t]
\begin{center}
   \centerline{\includegraphics[width=16cm]{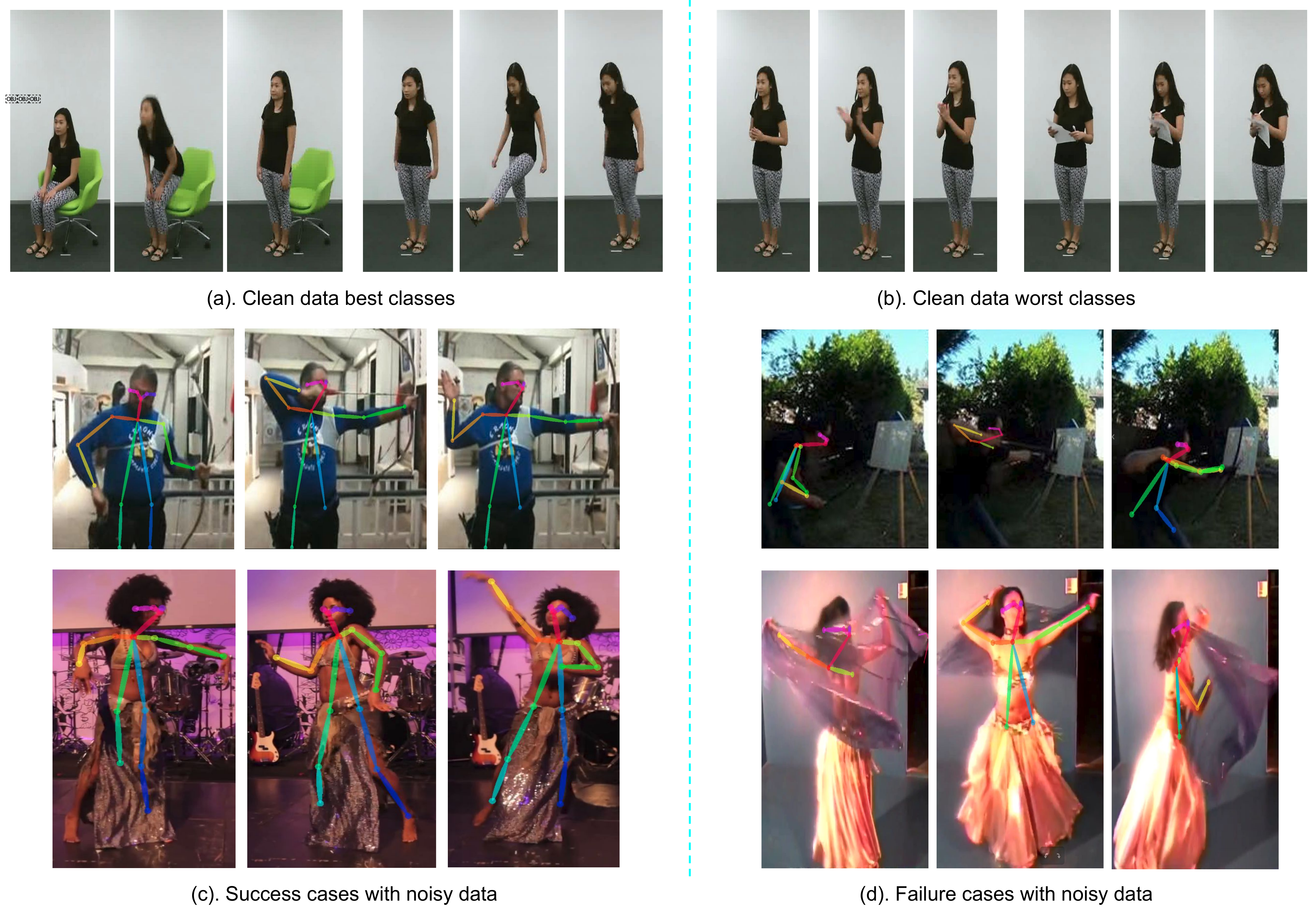}}
\end{center}
\vspace{-0.3in}
	\caption{Example frames from clean and noisy datasets. In the first row from left to right contains classes from NTU RGB+D: standing up, kicking something, clapping and writing. The second and third row contains predicted noisy poses of success and failure cases. The second row is from UCF101 and the third row is from Kinetics. Success cases are shown on the left side and the failure cases are shown on the right side.}
\label{fig:example_frame}
\end{figure*}

{\bf Kinetics Motion.}
As shown in Table \ref{table:kinetics_motion}, the proposed long term dependency model with attention is comparable to the state-of-the-art performances. The recognition accuracy also similar to the methods using other modalities including RGB and optical flow. This experiment proves that the proposed GLAN + SSAN framework is noise robust. Furthermore, it shows that skeleton based action recognition can achieve the state-of-the-art performance while requiring much less computations. The first, middle and last frames from example videos are shown for success and failure cases in the third row of Figure \ref{fig:example_frame}. Observations show that failure cases are mostly caused by missing or incorrect pose estimations. 
\begin{table}
\caption{The recognition accuracy compared to the state-of-the-art methods on the Kinetics Motion dataset.} 
\vspace{-0.15in}
\begin{center}
\begin{tabular}{ c c }
    \hline
    State-of-the-art & Accuracy (\%)\\
    \hline
	RGB CNN \cite{kay2017kinetics} & 70.4 \\
	Flow CNN \cite{kay2017kinetics} & 72.8 \\
	ST-GCN \cite{yan2018spatial} & 72.4 \\
    \hline
	Proposed Model & Accuracy (\%)\\
    \hline
	With TSSI & 58.8\\
	TSSI + GLAN \cite{yang2018action} & 67.2\\
	TSSI + SSAN + GLAN & 68.7\\
\end{tabular}
\end{center}
\vspace{-0.18in}
\label{table:kinetics_motion}
\end{table}

{\bf UCF101.} The model is also evaluated on the estimated poses from UCF101.  Compared to the state-of-the-art performance achieved by RGB + optical flow, the achieved accuracy of $43.5\%$ is not high. However similar to the idea of Kinetics Motion, we argue that the relatively low performance is because many classes in UCF101 require the information from objects and background scenes. Human poses alone are not sufficient for these classes, but are still of great importance in understanding human activities. The examples are shown in the second row of Figure \ref{fig:example_frame}.

\begin{table*}
\caption{The statistics and names of classes with the highest and lowest recognition accuracy. Experiments are conducted on NTU RGB+D with a cross subject setting. The lefthand table shows the classes that constantly have a good or bad performance. The righthand table shows the statistics of the top and bottom classes.} 
\vspace{-0.1in}
\begin{center}
\begin{tabular}{ccccc}
\hline
Selected Best Classes & Base. & TSSI & GLAN & GLAN + SSAN\\
\hline
standing up & 85.4 & 94.1 & {\bf 97.1} & 96.3 \\
sitting down & 91.6 & 91.6 & 93.8 & {\bf 93.8} \\
walking apart & 90.6 & 91.3 & 93.1 & {\bf 96.0} \\
kicking something & 80.8 & 91.7 & 92.4 & {\bf 92.8} \\
\hline
Selected Worst Classes & Base. & TSSI & GLAN & GLAN + SSAN\\
\hline
writing & {\bf 52.2} & 26.5 & 39.7 & 45.6 \\
reading & 25.6 & 26.0 & 39.9 & {\bf 42.8} \\
clapping & 17.2 & 36.6 & 39.7 & {\bf 63.0} \\
playing with phone & 31.6 & 43.6 & 56.0 & {\bf 66.2} \\
\hline
Overall & 68.0 & 73.1 & 80.1 & {\bf 82.4}\\
\hline
\end{tabular}
\quad
\begin{tabular}{ccccc}
\hline
Best Classes Stat. & Base. & TSSI & GLAN & GLAN + SSAN\\
\hline
Top 1 & 96.0 & {\bf 99.3} & 97.8 & 97.1\\
Top 3 Avg. & 93.6 & 96.1 & 96.8 & {\bf 96.8}\\
Top 5 Avg. & 92.0 & 94.3 & 96.2 & {\bf 96.6}\\
Top 10 Avg. & 87.3 & 92.0 & 94.8 & {\bf 95.5}\\
\hline
Worst Classes Stat. & Base. & TSSI & GLAN & GLAN + SSAN\\
\hline
Top 1 & 17.2 & 25.3 & 39.7 & {\bf 42.8}\\
Top 3 Avg. & 23.8 & 25.9 & 45.2 & {\bf 49.9}\\
Top 5 Avg. & 27.9 & 31.6 & 49.5 & {\bf 55.0}\\
Top 10 Avg. & 39.6 & 42.2 & 56.4 & {\bf 61.2}\\
\hline
Overall & 68.0 & 73.1 & 80.1 & {\bf 82.4}\\
\hline
\end{tabular}
\end{center}
\label{table:error_case}
\end{table*}

\section{Conclusions}
Using CNN for skeleton based action recognition is a promising approach. In this work, we address the two major problems with previous CNN based methods, i.e., the improper design of skeleton images and the lack of attention mechanisms. The design of skeleton images is improved by introducing the Tree Structure Skeleton Image (TSSI). The two-branch attention structure is then introduced for visual attention on the skeleton image. A Global Long-sequence Attention Network (GLAN) is proposed based on the two-branch attention structure. We further propose the long-term dependency model with a Sub-Sequence Attention Network (SSAN). The effectiveness of combining the GLAN and the SSAN is also validated. Experiments show that the proposed enhancement modules greatly improve the recognition accuracy, especially on the challenging classes. Furthermore, the model is robust on noisy estimated poses.


%



\section*{Acknowledgments}
The authors would like to thank the support of New York State through the Goergen Institute for Data Science, our corporate research sponsors Snap and Cheetah Mobile, and NSF Award \#1704309. 

\ifCLASSOPTIONcaptionsoff
  \newpage
\fi



%


\bibliographystyle{IEEEtran}
\bibliography{egbib.bib}

%

\begin{IEEEbiography}[{\includegraphics[width=1in,height=1.25in,clip]{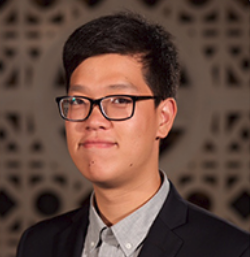}}]{Zhengyuan Yang}
received the BE degree in electrical engineering from the University of Science and Technology of China in 2016. He is currently pursuing the PhD degree with the Computer Science Department, University of Rochester, under the supervision of Prof. Jiebo Luo. His research interests mainly include action recognition, pose estimation and video analysis.
\end{IEEEbiography}

\begin{IEEEbiography}
[{\includegraphics[width=1in,height=1.25in,clip]{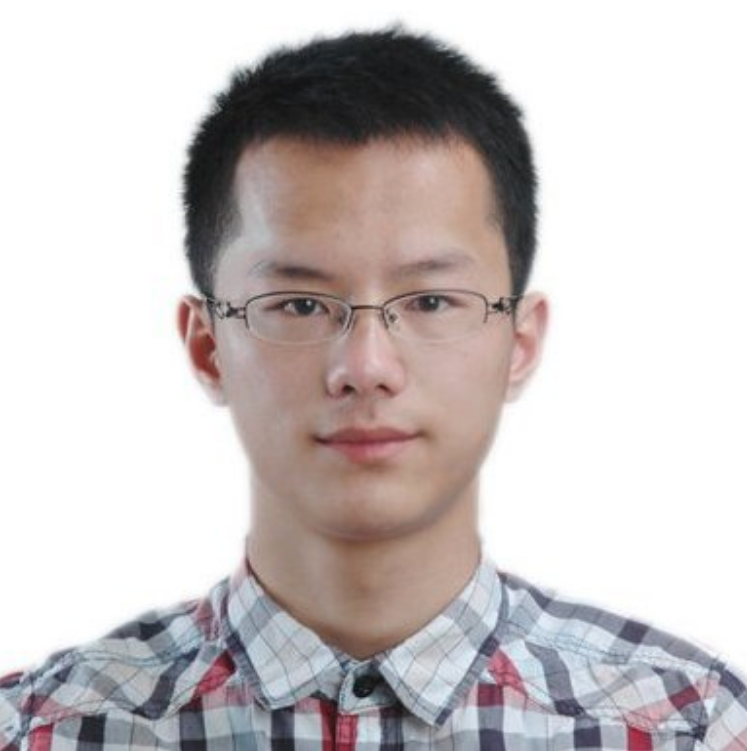}}]{Yuncheng Li}
received the BE degree in electrical engineering from the University of Science and Technology of China in 2012, and the PhD degree in computer sciences from the University of Rochester in 2017. He is currently a Research Scientist with Snapchat Inc.
\end{IEEEbiography}

\begin{IEEEbiography}
[{\includegraphics[width=1in,height=1.25in,clip]{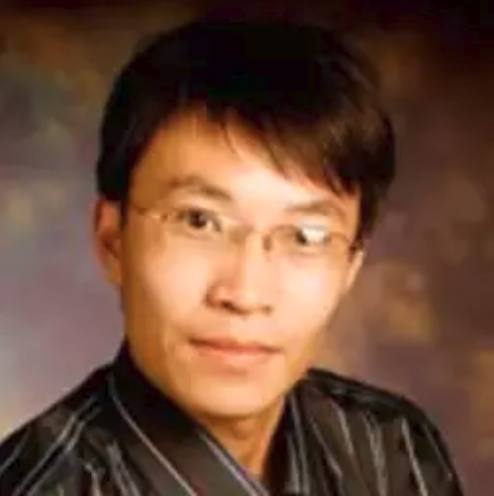}}]{Jianchao Yang}
(M'12) received the MS and PhD degrees from the ECE Department, University of Illinois at Urbana–Champaign, under the supervision of Prof. Thomas S. Huang. He was a Research Scientist with Adobe Research and is currently a Principal Research Scientist with Snapchat Inc. He has authored over 80 technical papers over a wide variety of topics in top tier conferences and journals, with over 12,000 citations per Google Scholar. His research focuses on computer vision, deep learning, and image and video processing. He received the Best Student Paper award in ICCV 2010, the Classification Task Prize in PASCAL VOC 2009, first position for object localization using external data in ILSVRC ImageNet 2014, and third place in the 2017 WebVision Challenge. He has served on the organizing committees of the ACM Multimedia Conference in 2017 and 2018.
\end{IEEEbiography}

\begin{IEEEbiography}
[{\includegraphics[width=1in,height=1.25in,clip]{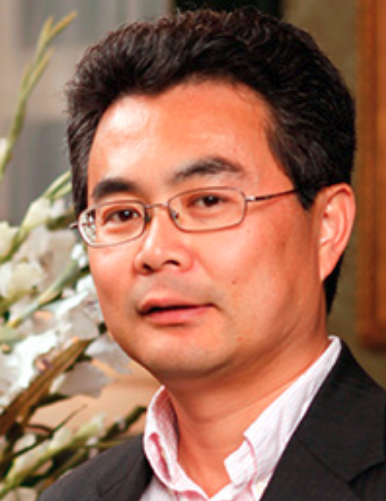}}]{Jiebo Luo}
(S'93-M'96-SM'99-F'09) joined the Department of Computer Science, University of Rochester, in 2011, after a prolific career of over 15 years with Kodak Research. He has authored over 300 technical papers and holds over 90 U.S. patents. His research interests include computer vision, machine learning, data mining, social media, and biomedical informatics. He has served as the Program Chair of the ACM Multimedia 2010,  IEEE CVPR 2012, ACM ICMR 2016, and IEEE ICIP 2017, and on the Editorial Boards of the IEEE TRANSACTIONS ON PATTERN ANALYSIS AND MACHINE INTELLIGENCE, IEEE TRANSACTIONS ON MULTIMEDIA, IEEE TRANSACTIONS ON CIRCUITS AND SYSTEMS FOR VIDEO TECHNOLOGY, Pattern Recognition, Machine Vision and Applications, and ACM Transactions on
Intelligent Systems and Technology. He is also a Fellow of the SPIE and IAPR.
\end{IEEEbiography}






\end{document}